\documentclass{bmvc2k}

\usepackage{amssymb}
\usepackage{bbm} % for \mathbbm
\usepackage{booktabs}       % Professional horizontal lines
\usepackage{multirow}
\usepackage{array}
\usepackage{tabularx}
\usepackage{tabulary}
\usepackage{tabularray}     % Modern advanced tables
\usepackage{ltablex}        % Longtable + Tabularx combined
\keepXColumns               % Essential for ltablex consistency
\usepackage{colortbl}       % Colored cells
\usepackage{threeparttable}
\usepackage{tablefootnote}
\usepackage{caption}
\usepackage{ragged2e}       % Better text alignment in cells
\usepackage{xcolor}

\title{MedCORE: Criteria-Grounded Clinical Reasoning for Interpretable Medical Image Diagnosis}

\addauthor{Asim Khan}
{asim.khan@ku.ac.ae}{1}
\addauthor{Samee Ullah Khan}
{samee.khan@ku.ac.ae}
{2}
\addauthor{Dwarikanath Mahapatra}{dwarikanath.mahapatra@ku.ac.ae}{1}

\addinstitution{
 Deparment of Computer Science, \\Khalifa University, Abu Dhabi, UAE
}
\addinstitution{
 Department of Aerosapce Engineering, \\Khalifa University, Abu Dhabi, UAE
}

\runninghead{Khan, Khan, Mahapatra}{MedCORE}

\def\eg{\emph{e.g}\bmvaOneDot}

\begin{document}

\maketitle

\begin{abstract}
Clinical diagnosis is inherently a structured reasoning process, yet existing deep learning models often bypass this structure by mapping image features directly to disease labels without explicitly interrogating the morphological and textural criteria that clinicians systematically evaluate. 
This limits diagnostic transparency and may compromise safe clinical deployment. 
We present MedCORE (Medical Criteria-Oriented Reasoning and Evidence), a structured diagnostic framework that operationalizes clinical reasoning within a vision-language architecture. 
For each input image, MedCORE decomposes the diagnostic process into clinically defined criteria, spatially localizes each criterion to diagnostically relevant image regions, encodes evidence through multi-scale representations that capture macro-structural and micro-textural pathological characteristics, and refines criterion representations using a Graph Attention Network that explicitly models inter-criteria dependencies.
Criterion representations are further aligned with clinical text descriptors, reinforced through class-wise visual prototypes, and aggregated using uncertainty-calibrated weighting that proportionally discounts low-confidence diagnostic evidence.
MedCORE is validated across three clinically heterogeneous imaging modalities, including dermoscopic lesion classification on ISIC 2018, breast ultrasound lesion characterization on BUSI, and diabetic retinopathy grading on IDRiD.
Quantitatively, MedCORE achieves 89.2\% accuracy, 85.7\% macro-F1, and 96.4\% AUC on ISIC 2018; 96.1\% accuracy, 95.2\% macro-F1, and 98.4\% AUC on BUSI; and 84.3\% accuracy, 80.2\% macro-F1, and 92.8\% AUC on IDRiD. 
These results demonstrate consistent improvements over strong CNN, transformer, biomedical vision-language, concept-based, and prototype-based baselines.
Collectively, the findings establish that grounding diagnostic inference in structured clinical criteria yields models that are accurate, interpretable, uncertainty-aware, and better aligned with the evidence-based reasoning used in clinical practice.
\end{abstract}

% %-------------------------------------------------------------------------
\section{Introduction}

Medical image diagnosis is increasingly supported by deep visual models, yet the dominant formulation still treats diagnosis as a direct mapping from pixels to disease labels. This formulation has delivered strong classification performance, but it does not match the way clinicians usually justify a decision. A dermatologist does not simply assign a lesion category; she inspects asymmetry, border irregularity, colour variation, dermoscopic structures, vascular patterns and other clinically meaningful cues. Likewise, ultrasound and retinal diagnosis rely on structured visual evidence that is spatially localised, clinically named and often mutually dependent. For high-stakes clinical deployment, a diagnostic model should therefore expose not only \emph{what} it predicts, but also \emph{which clinical criteria} support the prediction, \emph{where} those criteria are observed and \emph{how reliable} each piece of evidence is.

Vision--language pre-training offers a natural route toward this goal. General models such as CLIP \cite{radford2021learning} and biomedical variants including GLoRIA \cite{huang2021gloria}, BioViL \cite{boecking2022making}, MedKLIP \cite{wu2023medklip}, KAD \cite{zhang2023knowledge}, BiomedCLIP \cite{zhang2023biomedclip} and BiomedGPT \cite{zhang2024generalist} learn visual representations aligned with clinical or biomedical language. However, most of these methods rely on global image--text similarity, report-level supervision or weak region--word alignment. They may encode useful medical semantics, but the final classifier is often not forced to reason through explicit diagnostic criteria. Concept-based models address this limitation by inserting human-interpretable variables
between images and labels \cite{koh2020concept,yuksekgonul2022post, oikarinen2023label,yang2023language,srivastava2024vlg}. Recent medical approaches, including Explicd \cite{gao2024aligning}, MICA \cite{bie2024mica}, AdaCBM \cite{chowdhury2024adacbm} and Med-MICN \cite{hu2024towards}, show that clinical concepts can improve explanation quality. Yet three gaps remain. First, criteria are commonly represented as text anchors or concept tokens without explicit criterion-specific spatial grounding. Second, criteria are often treated independently, although clinical signs are correlated. Third, aggregation usually ignores criterion-level uncertainty, even when visual evidence is ambiguous or corrupted by imaging artefacts.

We propose \textbf{MedCORE}, a criteria-grounded framework for interpretable medical image classification. Instead of predicting labels directly from global image features, MedCORE decomposes diagnosis into clinically meaningful criteria. Each criterion is paired with a clinical text anchor encoded by the frozen BiomedCLIP text encoder, while a frozen BiomedCLIP visual backbone provides dense image tokens. Lightweight trainable modules then learn criterion-specific visual representations without updating the backbone. To make these representations spatially meaningful, learned criterion queries generate attention maps over image tokens, allowing each criterion to focus on diagnostically relevant regions without requiring spatial supervision. Multi-scale criterion encoding further captures both coarse morphological structure and fine local texture.

MedCORE further models diagnosis as structured evidence aggregation. A criterion dependency graph attention module \cite{velivckovic2017graph} refines each criterion using related criteria, allowing clinically correlated signs to reinforce one another. A
class-wise visual prototype memory maintains discriminative evidence for each criterion and disease class, encouraging compact same-class representations and better separation across classes. Finally, an uncertainty head estimates the reliability of each criterion and down-weights ambiguous evidence before the final prediction. The model is trained with a three-term objective: cross-entropy for diagnostic classification, vision--language alignment to clinical text anchors and prototype-based discriminative learning.

We evaluate MedCORE on ISIC2018 skin lesion classification \cite{codella2019skin}, BUSI breast ultrasound classification \cite{al2020dataset}, and diabetic-retinopathy grading using IDRiD/RetinaMNIST \cite{porwal2018idrid,yang2023medmnist}. These benchmarks span dermoscopy, ultrasound and retinal fundus imaging, enabling us to assess whether explicit clinical criteria, spatial grounding, relational reasoning, prototype reinforcement and uncertainty-calibrated aggregation improve both diagnostic performance and interpretability across heterogeneous imaging modalities.

Our main contributions are:
\begin{enumerate}
    \item We formulate medical image diagnosis as grounded clinical-criteria reasoning, enabling evidence-level interpretation across dermoscopy, breast ultrasound and retinal fundus imaging.

    \item We introduce criterion-aware spatial localisation with multi-scale criterion encoding, allowing each clinical criterion to attend to diagnostically relevant image regions and visual scales.

    \item We model inter-criteria dependencies using graph attention and reinforce class discriminative evidence through criterion-wise visual prototype learning.

    \item We propose uncertainty-calibrated criterion aggregation, reducing the influence of unreliable or ambiguous criteria in the final diagnostic decision.

    \item We evaluate MedCORE across three clinically heterogeneous medical image benchmarks using quantitative comparison, qualitative visualisation and ablation analysis.
\end{enumerate}

\section{Related Work}

\paragraph{Medical vision language learning.}
Large-scale vision language pre-training has become a strong foundation for medical image analysis. GLoRIA \cite{huang2021gloria} introduced global--local alignment between radiographs and reports for label-efficient recognition, while BioViL \cite{boecking2022making} exploited radiology text semantics to improve biomedical representation learning. MedKLIP \cite{wu2023medklip} and KAD \cite{zhang2023knowledge} inject medical knowledge into image--text pre-training, improving disease recognition and entity-level grounding in chest radiology. BiomedCLIP and BiomedGPT further extend biomedical vision language learning to broader image text corpora and diverse biomedical tasks \cite{zhang2023biomedclip,zhang2024generalist}. These works provide powerful backbones, but their explanations are typically derived from global similarity, report tokens or disease entities. MedCORE uses a biomedical vision--language backbone differently: clinical text anchors define explicit diagnostic criteria, while visual evidence is spatially grounded, relationally refined and uncertainty-weighted before classification.
\vspace{-5mm}
\paragraph{Concept-based interpretability and medical concept learning.}
Concept bottleneck models make predictions through human-understandable intermediate variables \cite{koh2020concept}. To reduce dependence on expensive concept annotations, recent post-hoc, label-free and language-guided CBMs use learned representations or language models to construct concept spaces \cite{yuksekgonul2022post,oikarinen2023label,yang2023language, srivastava2024vlg}. In medical imaging, Explicd aligns visual concept tokens with diagnostic knowledge, MICA performs multi-level image--concept alignment for skin lesions, AdaCBM adapts CLIP-based CBMs for diagnosis, and Med-MICN aligns concept semantics with neural-symbolic and saliency explanations \cite{gao2024aligning,bie2024mica,chowdhury2024adacbm,hu2024towards}. These methods show the value of clinical concepts, but most do not jointly address criterion-specific localisation, relationships among criteria and criterion-level uncertainty. MedCORE is designed around these three requirements.
\vspace{-5mm}
\paragraph{Spatial grounding, relational reasoning and uncertainty.}
Spatial grounding is particularly important in medicine because many diagnostic criteria are valid only in specific regions. Segmentation and foundation models such as SAM and MedSAM demonstrate the value of spatial priors in natural and medical image analysis \cite{kirillov2023segment,ma2024segment}. However, diagnostic classification often requires criterion-specific evidence rather than generic masks or image-level saliency. MedCORE therefore learns criterion-specific attention maps directly from dense visual tokens, localising evidence for each clinical criterion without requiring pixel-level criterion annotations. Clinical criteria are also relational: for example, asymmetry, border irregularity and colour variation may provide mutually reinforcing evidence in dermoscopy, while microaneurysms, haemorrhages and exudates jointly inform diabetic retinopathy severity. Graph attention networks provide a compact mechanism for learning such dependencies \cite{velivckovic2017graph}. Prototype-based models such as PIP-Net and ProtoViT improve interpretability through visual prototypes \cite{nauta2023pipnet,ma2024interpretable}; MedCORE instead uses class-wise prototypes as
criterion-level discriminative memory while keeping explanations organised around clinical criteria. Finally, recent work emphasizes the importance of uncertainty in medical image classification \cite{chen2024modeling}. Rather than estimating only image-level confidence, MedCORE assigns uncertainty to each criterion and uses it to calibrate evidence aggregation.
%------------------------

\section{Methodology}
\label{sec:method}

\begin{figure*}[!t]
\centering
\includegraphics[width=\textwidth]{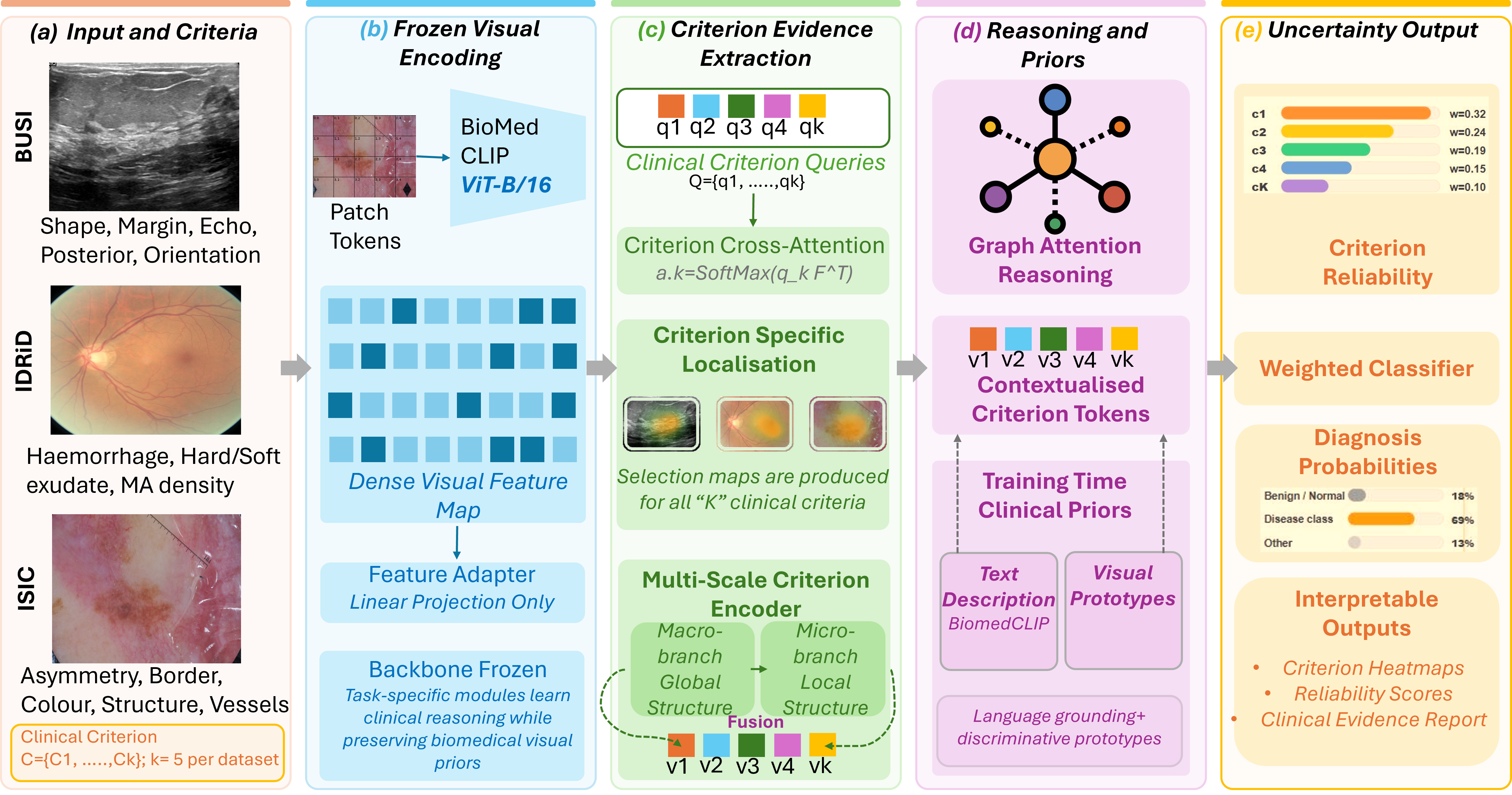}
\vspace{0.4em}
\caption{Clinically guided interpretable medical image classification framework. Input images are encoded using a frozen BioMedCLIP ViT-B/16 backbone, while clinical criterion queries extract localized evidence. The extracted evidence is refined through multi-scale encoding and graph attention reasoning with clinical priors. The model outputs diagnosis probabilities, criterion reliability scores, uncertainty estimates, and interpretable heatmaps.}
\label{fig:block}
\vspace{-5mm}
\end{figure*}

\subsection{Problem Formulation}
\label{sec:problem}

Let \(\mathbf{I} \in \mathbb{R}^{H \times W \times 3}\) denote a medical image and $y \in \{1, \ldots, C\}$ its ground-truth disease label.
Rather than learning a direct mapping $f: \mathbf{I} \mapsto y$, MedCORE decomposes the diagnostic decision into $K$ dataset-specific \emph{clinical criteria} $\{c_k\}_{k=1}^{K}$, each representing a morphological or textural property that clinicians systematically evaluate (\eg, asymmetry and border irregularity in dermoscopy; shape and margin in breast ultrasound; microaneurysms and haemorrhages in diabetic retinopathy grading).
The model evaluates each criterion independently, explicitly models inter-criteria dependencies, and aggregates evidence into a final diagnosis proportionally weighted by per-criterion confidence.

\subsection{Visual Backbone and Feature Extraction}
\label{sec:backbone}

Given input image $\mathbf{I}$, a shared visual backbone (\eg, BiomedCLIP ViT-B/16~\cite{zhang2023biomedclip}) produces a dense feature map:
\begin{equation}
    \mathbf{F} \in \mathbb{R}^{N_p \times D},
    \label{eq:backbone}
\end{equation}
where $N_p = H'W'$ is the number of spatial tokens and $D$ is the embedding dimension.
The backbone is frozen; only task-specific modules introduced below are trained, preserving the rich biomedical visual priors acquired
during pre-training.

\subsection{Criterion-Aware Spatial Localisation}
\label{sec:localise}

Different clinical criteria manifest in distinct anatomical regions of the image. For criterion $k$, we learn a spatial attention map
$\mathbf{a}_k \in \mathbb{R}^{N_p}$ via a lightweight query-key mechanism:
\begin{equation}
    \mathbf{a}_k = \mathrm{softmax}\!\left(
        \frac{\mathbf{q}_k \mathbf{F}^{\top}}{\sqrt{D}}
    \right), \quad
    \mathbf{r}_k = \mathbf{a}_k \mathbf{F},
    \label{eq:localise}
\end{equation}
where $\mathbf{q}_k \in \mathbb{R}^{1 \times D}$ is a learnable criterion query vector and $\mathbf{r}_k \in \mathbb{R}^{D}$ is the resulting criterion-specific region descriptor.
This produces $K$ spatially grounded representations, one per criterion, without requiring spatial supervision.

\subsection{Multi-Scale Criterion Encoding}
\label{sec:multiscale}

Pathological features operate at two complementary scales: macro-structural (\eg, overall lesion shape, global tissue organisation) and micro-textural (\eg, fine-grained surface patterns, local intensity heterogeneity).
For criterion $k$ we encode both scales separately and fuse them:
\begin{equation}
    \mathbf{v}_k = \mathbf{W}_{\mathrm{fuse}}\,
    \bigl[\,\mathbf{v}_k^{\mathrm{macro}};\,\mathbf{v}_k^{\mathrm{micro}}\,\bigr]
    + \mathbf{b},
    \label{eq:multiscale}
\end{equation}
where $\mathbf{v}_k^{\mathrm{macro}}$ is obtained by global average pooling over $\mathbf{r}_k$ and a linear projection, $\mathbf{v}_k^{\mathrm{micro}}$ is derived from a local patch neighbourhood of the peak attention location in $\mathbf{a}_k$ passed through a depthwise convolutional encoder, $[\,;\,]$ denotes concatenation, and $\mathbf{W}_{\mathrm{fuse}} \in \mathbb{R}^{D \times 2D}$ is a learnable fusion matrix.
The resulting $\mathbf{v}_k \in \mathbb{R}^{D}$ encodes the criterion's evidence at both scales.

\subsection{Graph Attention Network for Inter-Criteria Reasoning}
\label{sec:gat}

Clinical criteria are not independent: a finding of malignant asymmetry strengthens the evidential weight of irregular borders, and the co-presence of hard exudates and microaneurysms jointly elevates the DR severity grade.
MedCORE models these dependencies explicitly by constructing a fully connected criterion graph $\mathcal{G} = (\mathcal{V}, \mathcal{E})$, where nodes $\mathcal{V} = \{\mathbf{v}_k\}_{k=1}^{K}$ are criterion representations and edges $\mathcal{E}$ carry learnable relational weights.

A single Graph Attention Network (GAT)~\cite{velivckovic2017graph} layer refines each node by attending over its neighbours:
\begin{align}
    e_{kj} &= \mathrm{LeakyReLU}\!\left(
        \mathbf{a}^{\top}\bigl[\mathbf{W}_g \mathbf{v}_k \,\|\,
        \mathbf{W}_g \mathbf{v}_j\bigr]
    \right),
    \label{eq:gat_attn} \\
    \alpha_{kj} &= \frac{\exp(e_{kj})}
        {\sum_{j' \in \mathcal{N}(k)} \exp(e_{kj'})},
    \label{eq:gat_norm} \\
    \tilde{\mathbf{v}}_k &= \sigma\!\left(
        \sum_{j \in \mathcal{N}(k)} \alpha_{kj}\,\mathbf{W}_g\,\mathbf{v}_j
    \right),
    \label{eq:gat_out}
\end{align}
where $\mathbf{W}_g \in \mathbb{R}^{D' \times D}$ is a shared linear projection, $\mathbf{a} \in \mathbb{R}^{2D'}$ is the attention vector, $\mathcal{N}(k)$ denotes all criteria (full graph), $\|$ is vector concatenation, and $\sigma$ is ELU.
Multi-head attention ($H=4$ heads) is used and outputs are concatenated.
The refined representations $\{\tilde{\mathbf{v}}_k\}_{k=1}^{K}$ encode each criterion in the context of all others.

\subsection{Vision–Language Criterion Alignment}
\label{sec:vl}

Each criterion $c_k$ is paired with a short clinical text descriptor (\eg, \emph{``irregular, asymmetric border with notching''} for dermoscopic border criterion).
The frozen BiomedCLIP text encoder maps this to a text embedding $\mathbf{t}_k \in \mathbb{R}^{D}$.
We align $\tilde{\mathbf{v}}_k$ and $\mathbf{t}_k$ using a symmetric InfoNCE contrastive loss over a batch of $B$ samples:
\begin{equation}
    \mathcal{L}_{\mathrm{VL}} = -\frac{1}{K} \sum_{k=1}^{K}
    \log \frac{
        \exp\!\left(\tilde{\mathbf{v}}_k^{\top} \mathbf{t}_k \,/\, \tau\right)
    }{
        \sum_{j=1}^{K} \exp\!\left(\tilde{\mathbf{v}}_k^{\top} \mathbf{t}_j \,/\, \tau\right)
    },
    \label{eq:vl_loss}
\end{equation}
where $\tau$ is a learnable temperature.
This loss pulls each criterion's visual representation towards its corresponding clinical description while pushing it away from the descriptions of other criteria, grounding visual features in clinically meaningful language.

\subsection{Class-Wise Visual Prototype Reinforcement}
\label{sec:proto}

To further sharpen criterion discriminability, we maintain a set of class-wise visual prototypes $\mathbf{P}_k = \{\mathbf{p}_{k,c}\}_{c=1}^{C} \subset \mathbb{R}^{D}$ for each criterion $k$.
These prototypes are updated online as exponential moving averages of the criterion features from each class.
The prototype prediction for criterion $k$ is computed as the cosine similarity to each class prototype:
\begin{equation}
    \hat{p}_{k,c} = \frac{
        \exp\!\left(\mathrm{sim}(\tilde{\mathbf{v}}_k,\,\mathbf{p}_{k,c})\,/\,\tau_p\right)
    }{
        \sum_{c'=1}^{C}
        \exp\!\left(\mathrm{sim}(\tilde{\mathbf{v}}_k,\,\mathbf{p}_{k,c'})\,/\,\tau_p\right)
    },
    \label{eq:proto}
\end{equation}
where $\mathrm{sim}(\mathbf{u}, \mathbf{v}) = \mathbf{u}^{\top}\mathbf{v} /
(\|\mathbf{u}\|\|\mathbf{v}\|)$ is cosine similarity and $\tau_p$ is a fixed temperature.
The prototype cross-entropy reinforces class-discriminative structure within each criterion's feature space:
\begin{equation}
    \mathcal{L}_{\mathrm{proto}} = -\frac{1}{K}\sum_{k=1}^{K}
    \log \hat{p}_{k,y},
    \label{eq:proto_loss}
\end{equation}
where $y$ is the ground-truth class label.

\subsection{Uncertainty-Calibrated Criterion Aggregation}
\label{sec:uncertainty}

Not all criteria are equally informative for every sample.
For ambiguous cases, a criterion may produce unreliable evidence that should contribute less to the final decision.
For each criterion $k$, MedCORE predicts a scalar uncertainty estimate $\hat{u}_k \in \mathbb{R}^{+}$ via a two-layer MLP applied to $\tilde{\mathbf{v}}_k$.
Criterion contributions to the final class logits are then weighted inversely by their uncertainty through a softmax-normalised scheme:
\begin{align}
    w_k &= \frac{\exp(-\hat{u}_k)}
               {\sum_{j=1}^{K} \exp(-\hat{u}_j)},
    \label{eq:unc_weight} \\[6pt]
    \hat{\mathbf{y}} &= \sum_{k=1}^{K} w_k\,\mathbf{W}_{\mathrm{cls}}\,\tilde{\mathbf{v}}_k,
    \label{eq:aggregation}
\end{align}
where $\mathbf{W}_{\mathrm{cls}} \in \mathbb{R}^{C \times D}$ is a shared classification head.
High uncertainty $\hat{u}_k$ suppresses criterion $k$'s vote; a criterion that fires confidently on diagnostically decisive features
dominates the aggregation.

\subsection{Training Objective}
\label{sec:loss}

MedCORE is trained end-to-end with a three-term objective:
\begin{equation}
    \mathcal{L} = \mathcal{L}_{\mathrm{CE}}
    + \lambda_{\mathrm{VL}}\,\mathcal{L}_{\mathrm{VL}}
    + \lambda_{\mathrm{proto}}\,\mathcal{L}_{\mathrm{proto}},
    \label{eq:total_loss}
\end{equation}
where $\mathcal{L}_{\mathrm{CE}}$ is the standard cross-entropy loss on the aggregated logits $\hat{\mathbf{y}}$, $\mathcal{L} {\mathrm{VL}}$ (Eq.~\ref{eq:vl_loss}) enforces language alignment, $\mathcal{L}_{\mathrm{proto}}$ (Eq.~\ref{eq:proto_loss}) reinforces prototype discriminability, and $\lambda_{\mathrm{VL}}, \lambda_{\mathrm{proto}} > 0$ are balancing coefficients.
The uncertainty head is trained implicitly: criteria that consistently mislead the classification loss are penalised through the gradient
signal, causing the MLP to learn higher $\hat{u}_k$ for unreliable criteria without requiring explicit uncertainty supervision.

% \subsection{Clinical Criteria per Dataset}
% \label{sec:criteria}

%----------------------------
\section{Experimental Analysis}
\paragraph{Datasets.}
We evaluate MedCORE on three public medical image benchmarks spanning dermoscopy, breast ultrasound, and retinal fundus imaging. These datasets cover clinically distinct diagnostic settings and allow us to test whether criteria-grounded reasoning generalizes across lesion morphology, sonographic appearance, and retinal pathology.
% \paragraph{Dermoscopy (ISIC 2018) \cite{codella2019skin}.}
ISIC 2018 \cite{codella2019skin} is a dermoscopic skin lesion dataset designed for automated skin lesion analysis and melanoma-related diagnosis. It contains lesion images with diagnostic categories and is well suited for criteria such as asymmetry, border irregularity, colour variation, dermoscopic structures, and vascular patterns, which are central to dermatological visual assessment.
% \paragraph{Breast Ultrasound (BUSI) \cite{al2020dataset}.}
BUSI \cite{al2020dataset} is a breast ultrasound dataset containing normal, benign, and malignant cases. It captures grayscale sonographic appearances of breast tissue and lesions, making it suitable for BI-RADS-inspired criteria such as lesion shape, margin characteristics, internal echo pattern, posterior acoustic features, and lesion orientation.
% \paragraph{Diabetic Retinopathy (IDRiD) \cite{porwal2018idrid}.}
IDRiD \cite{porwal2018idrid} is a retinal fundus image dataset for diabetic retinopathy analysis. It
provides image-level disease severity information and lesion annotations for typical diabetic retinopathy findings, making it appropriate for ETDRS-aligned criteria such as microaneurysms, haemorrhages, hard exudates, soft exudates, and neovascularisation.

We evaluate MedCORE on three representative medical image classification tasks covering different imaging modalities and diagnostic criteria: dermoscopic lesion classification, breast ultrasound classification, and diabetic retinopathy grading. The evaluated datasets are ISIC 2018~\cite{codella2019skin}, BUSI~\cite{al2020dataset}, and IDRiD~\cite{porwal2018idrid}. These datasets were selected because their diagnostic processes naturally depend on structured clinical criteria rather than a single global image pattern. Table~\ref{tab:dataset_setup} summarizes the task definitions and the clinical criteria used by MedCORE for each dataset.

All images are resized to $224\times224$ and normalized before being passed to the visual backbone. During training, we apply standard image augmentations, including random horizontal flipping, random resized cropping, mild rotation, and color jittering where appropriate. The same preprocessing and data partitions are used for all competing methods to ensure a fair comparison. For MedCORE, the visual backbone is kept frozen, and only the criterion queries, multi-scale criterion encoder, graph reasoning module, uncertainty head, prototype module, and classifier are optimized.

Unless otherwise specified, MedCORE uses $K=5$ clinical criteria for each dataset. The criterion graph is fully connected, and the graph attention module uses four attention heads. The model is trained using AdamW with a learning rate of $1\times10^{-4}$ and weight decay of $1\times10^{-4}$. The batch size is set to 32, and training is performed for 100 epochs with early stopping based on validation Macro-F1. The loss balancing weights are set to $\lambda_{\mathrm{VL}}=0.1$ and $\lambda_{\mathrm{proto}}=0.1$. The prototype temperature is fixed to $\tau_p=0.07$, while the vision--language temperature $\tau$ is learnable. All results are reported on the held-out test set.

\begin{table*}[t]
\centering
\caption{Experimental setup and dataset-specific clinical criteria used by MedCORE.}
\label{tab:dataset_setup}
\small
\setlength{\tabcolsep}{4pt}
\begin{tabularx}{\textwidth}{@{}p{1.6cm}p{3.4cm}p{1.8cm}X@{}}
\toprule
\textbf{Dataset} & \textbf{Task} & \textbf{Classes} & \textbf{Clinical criteria used by MedCORE} \\
\midrule
ISIC 2018 & Dermoscopic lesion classification & 7 classes &
Asymmetry; border irregularity; colour variation; dermoscopic structures; vascular patterns \\

BUSI & Breast ultrasound classification & 3 classes &
Lesion shape; margin characteristics; internal echo pattern; posterior acoustic features; lesion orientation \\

IDRiD & Diabetic retinopathy grading & 5 grades &
Microaneurysm density; haemorrhage severity; hard exudate extent; soft exudate presence; neovascularisation \\
\bottomrule
\end{tabularx}
\end{table*}

\subsection{Evaluation Metrics}
\label{sec:evaluation_metrics}

We evaluate classification performance using accuracy, Macro-F1, and area under the receiver operating characteristic curve (AUC). Accuracy measures the overall proportion of correctly classified test samples:
\begin{equation}
    \mathrm{Accuracy}=\frac{1}{N}\sum_{i=1}^{N}\mathbbm{1}(\hat{y}_i=y_i),
\end{equation}
where $N$ is the number of test samples, $y_i$ is the ground-truth label, and $\hat{y}_i$ is the predicted label. Since the evaluated medical datasets are class-imbalanced, Macro-F1 is used as the primary metric. For each class $c$, precision, recall, and F1-score are computed as
\begin{equation}
    \mathrm{Precision}_{c}=\frac{\mathrm{TP}_{c}}{\mathrm{TP}_{c}+\mathrm{FP}_{c}}, \quad
    \mathrm{Recall}_{c}=\frac{\mathrm{TP}_{c}}{\mathrm{TP}_{c}+\mathrm{FN}_{c}},
\end{equation}
\begin{equation}
    \mathrm{F1}_{c}=\frac{2\cdot \mathrm{Precision}_{c}\cdot \mathrm{Recall}_{c}}
    {\mathrm{Precision}_{c}+\mathrm{Recall}_{c}}, \quad
    \mathrm{Macro\text{-}F1}=\frac{1}{C}\sum_{c=1}^{C}\mathrm{F1}_{c}.
\end{equation}
AUC is computed using one-vs-rest classification for multi-class tasks and then averaged across classes.

In addition to discrimination performance, we evaluate calibration using reliability diagrams. For a set of confidence bins $\{B_m\}_{m=1}^{M}$, the expected calibration error (ECE) is defined as
\begin{equation}
    \mathrm{ECE}=\sum_{m=1}^{M}\frac{|B_m|}{N}
    \left|\mathrm{acc}(B_m)-\mathrm{conf}(B_m)\right|,
\end{equation}
where $\mathrm{acc}(B_m)$ is the empirical accuracy of samples in bin $B_m$, and $\mathrm{conf}(B_m)$ is their average predicted confidence. A well-calibrated model has empirical accuracy close to confidence across all bins.

\subsection{Comparison with Baseline Methods}
\label{sec:baseline_comparison}

We compare MedCORE with representative CNN, transformer, biomedical vision--language, concept-based, and prototype-based baselines. CNN baselines include DenseNet--121~\cite{huang2017densely}, ResNet-50~\cite{koonce2021resnet}, and EfficientNet-B0~\cite{tan2019efficientnet}. Transformer baselines include ViT-B/16~\cite{dosovitskiy2021image} and Swin-T~\cite{liu2021swin}. To assess the benefit of clinical-criterion reasoning beyond biomedical pretraining alone, we include BiomedCLIP \cite{zhang2023biomedclip} with linear probing and fine-tuning. We also compare with Concept Bottleneck Models~\cite{koh2020concept} and ProtoPNet~\cite{chen2019looks}, which provide interpretable intermediate concepts or prototypes.

Table~\ref{tab:main_results} reports the complete classification results. MedCORE achieves the best performance across all three datasets, obtaining an average Macro-F1 of 87.0\%. Compared with BiomedCLIP fine-tuning, MedCORE improves Macro-F1 by 3.3 percentage points on ISIC 2018, 2.4 percentage points on BUSI, and 3.7 percentage points on IDRiD. These gains indicate that the improvement does not come only from using a strong biomedical visual backbone; rather, explicit clinical-criterion localization, inter-criterion reasoning, and uncertainty-calibrated aggregation provide additional benefits.

\begin{table*}[!t]
\centering
\caption{Quantitative comparison with baseline methods. Accuracy, Macro-F1, and AUC are reported as percentages. The last column reports the average Macro-F1 across the three datasets.}
\vspace{1mm}
\label{tab:main_results}
\vspace{0.4em}
\small
\setlength{\tabcolsep}{2pt}
\begin{tabular*}{\textwidth}{@{\extracolsep{\fill}}lcccccccccc@{}}
\toprule
\multirow{2}{*}{\textbf{Method}} 
& \multicolumn{3}{c}{\textbf{ISIC 2018}} 
& \multicolumn{3}{c}{\textbf{BUSI}} 
& \multicolumn{3}{c}{\textbf{IDRiD}} 
& \multirow{2}{*}{\textbf{Avg. F1}} \\
\cmidrule(lr){2-4}
\cmidrule(lr){5-7}
\cmidrule(lr){8-10}
& \textbf{Acc.} & \textbf{F1} & \textbf{AUC} 
& \textbf{Acc.} & \textbf{F1} & \textbf{AUC} 
& \textbf{Acc.} & \textbf{F1} & \textbf{AUC} 
& \\
\midrule
ResNet-50 \cite{koonce2021resnet} & 81.2 & 75.8 & 90.4 & 87.4 & 85.9 & 93.5 & 73.5 & 68.4 & 84.7 & 76.7 \\
DenseNet-121~\cite{huang2017densely} & 82.7 & 77.3 & 91.2 & 89.2 & 87.8 & 94.6 & 75.4 & 70.1 & 86.3 & 78.4 \\
EfficientNet-B0 \cite{tan2019efficientnet} & 84.1 & 78.8 & 92.0 & 90.6 & 89.1 & 95.5 & 76.9 & 72.0 & 87.5 & 80.0 \\
ViT-B/16~\cite{dosovitskiy2021image} & 84.8 & 79.6 & 92.5 & 91.2 & 89.8 & 96.1 & 78.0 & 73.1 & 88.4 & 80.8 \\
Swin-T~\cite{liu2021swin}  & 85.6 & 80.8 & 93.1 & 92.1 & 90.7 & 96.6 & 79.2 & 74.6 & 89.2 & 82.0 \\
BioMedCLIP-LP \cite{zhang2023biomedclip} & 85.9 & 81.2 & 93.8 & 93.3 & 91.7 & 97.0 & 79.6 & 75.0 & 89.7 & 82.6 \\
BioMedCLIP-FT \cite{zhang2023biomedclip} & 87.0 & 82.4 & 94.7 & 94.2 & 92.8 & 97.5 & 81.0 & 76.5 & 90.8 & 83.9 \\
Concept Bottleneck Model \cite{koh2020concept} & 85.5 & 80.7 & 93.2 & 92.4 & 91.1 & 96.6 & 79.5 & 74.4 & 89.2 & 82.1 \\
ProtoPNet~\cite{chen2019looks} & 86.1 & 81.5 & 94.1 & 93.0 & 91.8 & 97.0 & 80.1 & 75.7 & 90.0 & 83.0 \\
\textbf{MedCORE} 
& \textbf{89.2} & \textbf{85.7} & \textbf{96.4} 
& \textbf{96.1} & \textbf{95.2} & \textbf{98.4} 
& \textbf{84.3} & \textbf{80.2} & \textbf{92.8} 
& \textbf{87.0} \\
\bottomrule
\end{tabular*}
\end{table*}

The largest absolute improvement is observed on IDRiD, where grading diabetic retinopathy depends on multiple interacting lesion types. This supports the motivation for graph-based inter-criterion reasoning. The improvement on ISIC 2018 also shows that decomposing lesion diagnosis into ABCD-inspired criteria provides stronger supervision than global visual classification. BUSI obtains the highest absolute scores, suggesting that the BI-RADS-inspired criterion set captures diagnostically meaningful ultrasound patterns.

\subsection{Ablation Study}
\label{sec:ablation_study}

\begin{figure}[!t]
    \centering
    \includegraphics[width=\linewidth]{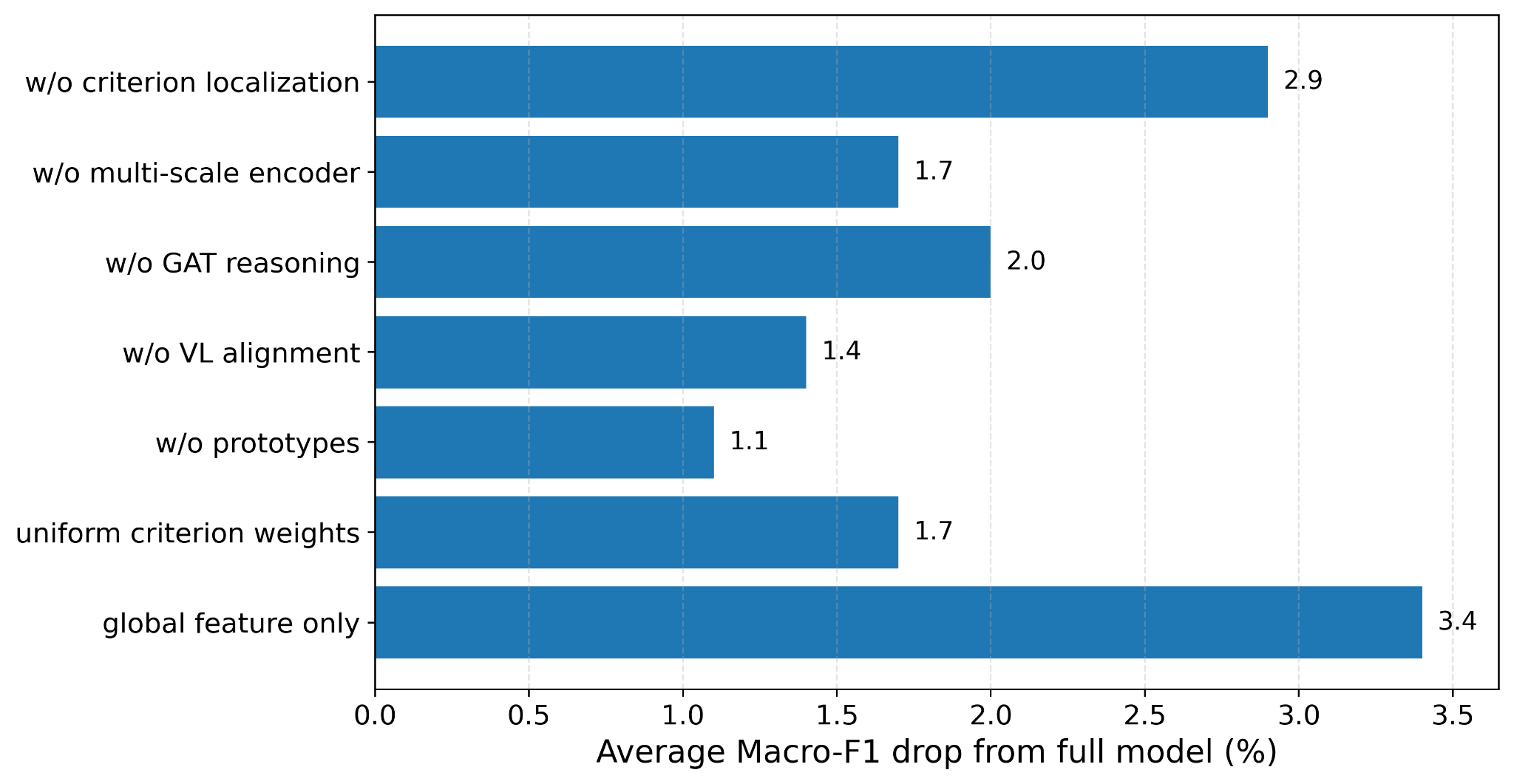}
    \caption{Ablation study showing the average Macro-F1 drop from the full MedCORE model. Larger values indicate a more important component.}
    \label{fig:ablation_given}
\end{figure}

We conduct ablation experiments to quantify the contribution of each component in MedCORE. Figure~\ref{fig:ablation_given} reports the average Macro-F1 drop after removing individual modules, and Table~\ref{tab:ablation_results} provides the corresponding values.

\begin{table*}[!t]
\centering
\caption{Ablation study based on the average Macro-F1 over ISIC 2018, BUSI, and IDRiD. The full MedCORE model obtains an average Macro-F1 of 87.0\%.}
\label{tab:ablation_results}
\vspace{1mm}
\small
\setlength{\tabcolsep}{6pt}
\begin{tabular*}{\textwidth}{@{\extracolsep{\fill}}lcc@{}}
\toprule
\textbf{Variant} & \textbf{F1 drop} & \textbf{Avg. F1} \\
\midrule
\textbf{Full MedCORE} & \textbf{0.0} & \textbf{87.0} \\
w/o criterion localization & 2.9 & 84.1 \\
w/o multi-scale encoder & 1.7 & 85.3 \\
w/o GAT reasoning & 2.0 & 85.0 \\
w/o VL alignment & 1.4 & 85.6 \\
w/o prototypes & 1.1 & 85.9 \\
uniform criterion weights & 1.7 & 85.3 \\
global feature only & 3.4 & 83.6 \\
\bottomrule
\end{tabular*}
\end{table*}

The largest degradation occurs when replacing criterion-level reasoning with a single global feature representation, which reduces average Macro-F1 by 3.4 percentage points. This confirms that the proposed clinical decomposition is central to MedCORE. Removing criterion localization produces the second largest drop of 2.9 percentage points, showing that different clinical criteria require spatially distinct evidence. Removing GAT reasoning decreases Macro-F1 by 2.0 percentage points, indicating that dependencies among clinical findings are useful for diagnosis. The multi-scale encoder and uncertainty weighting each contribute 1.7 percentage points, demonstrating the importance of combining macro-structural and micro-textural evidence and suppressing unreliable criterion predictions. Vision--language alignment and prototype reinforcement provide smaller but consistent gains, suggesting that clinical semantic grounding and class-wise visual prototypes regularize the criterion feature space.

\subsection{Sensitivity to the Number of Clinical Criteria}
\label{sec:k_sensitivity}

\begin{figure}[!t]
    \centering
    \includegraphics[width=0.7\linewidth]{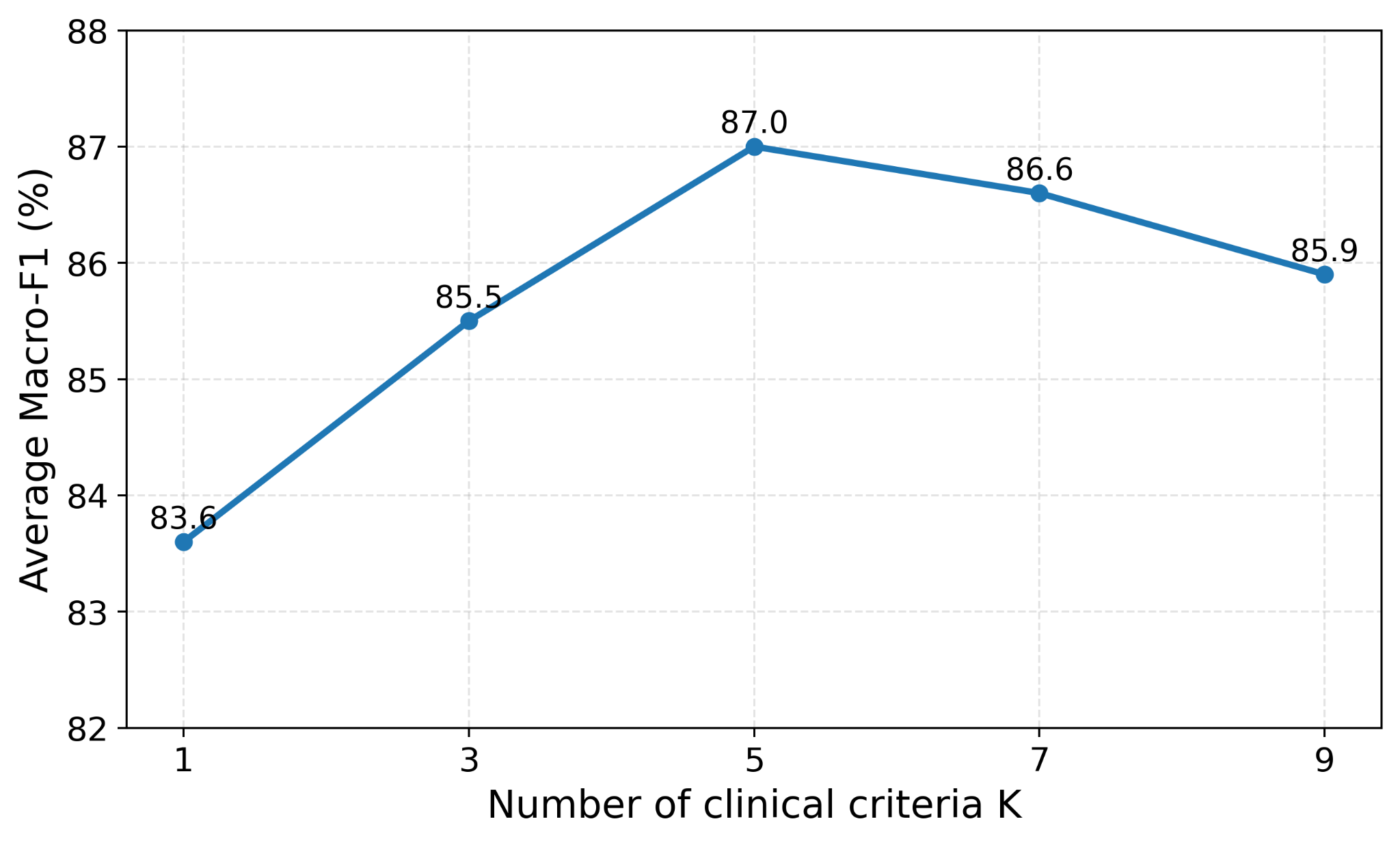}
    \caption{Sensitivity analysis with respect to the number of clinical criteria $K$. The best performance is obtained with $K=5$, which corresponds to the clinically defined criterion set used in MedCORE.}
    \label{fig:k_sensitivity_given}
\end{figure}

Figure~\ref{fig:k_sensitivity_given} and Table~\ref{tab:k_sensitivity} show the effect of varying the number of clinical criteria $K$. The average Macro-F1 improves from 83.6\% at $K=1$ to 87.0\% at $K=5$, and then gradually decreases when $K$ is increased to 7 or 9.

\begin{table}[!t]
\centering
\caption{Average Macro-F1 under different numbers of clinical criteria.}
\label{tab:k_sensitivity}
\vspace{1mm}
\small
\setlength{\tabcolsep}{8pt}
\begin{tabular}{@{}cc@{}}
\toprule
\textbf{Number of criteria $K$} & \textbf{Avg. Macro-F1} \\
\midrule
1 & 83.6 \\
3 & 85.5 \\
5 & \textbf{87.0} \\
7 & 86.6 \\
9 & 85.9 \\
\bottomrule
\end{tabular}
\end{table}

The poor performance at $K=1$ indicates that a single global representation is insufficient to capture diverse diagnostic evidence. Increasing $K$ to 3 improves performance because the model can separate multiple clinical cues. The best result is obtained at $K=5$, which matches the manually defined clinical criterion set for each dataset. Increasing $K$ beyond 5 slightly reduces performance, likely because additional criteria become redundant, weakly defined, or noisy, making criterion localization and graph reasoning less stable. This result supports the use of a compact clinically motivated criterion set rather than an excessively large number of latent concepts.

\subsection{Calibration Analysis}
\label{sec:calibration_analysis}

\begin{figure}[!t]
    \centering
    \includegraphics[width=0.7\linewidth]{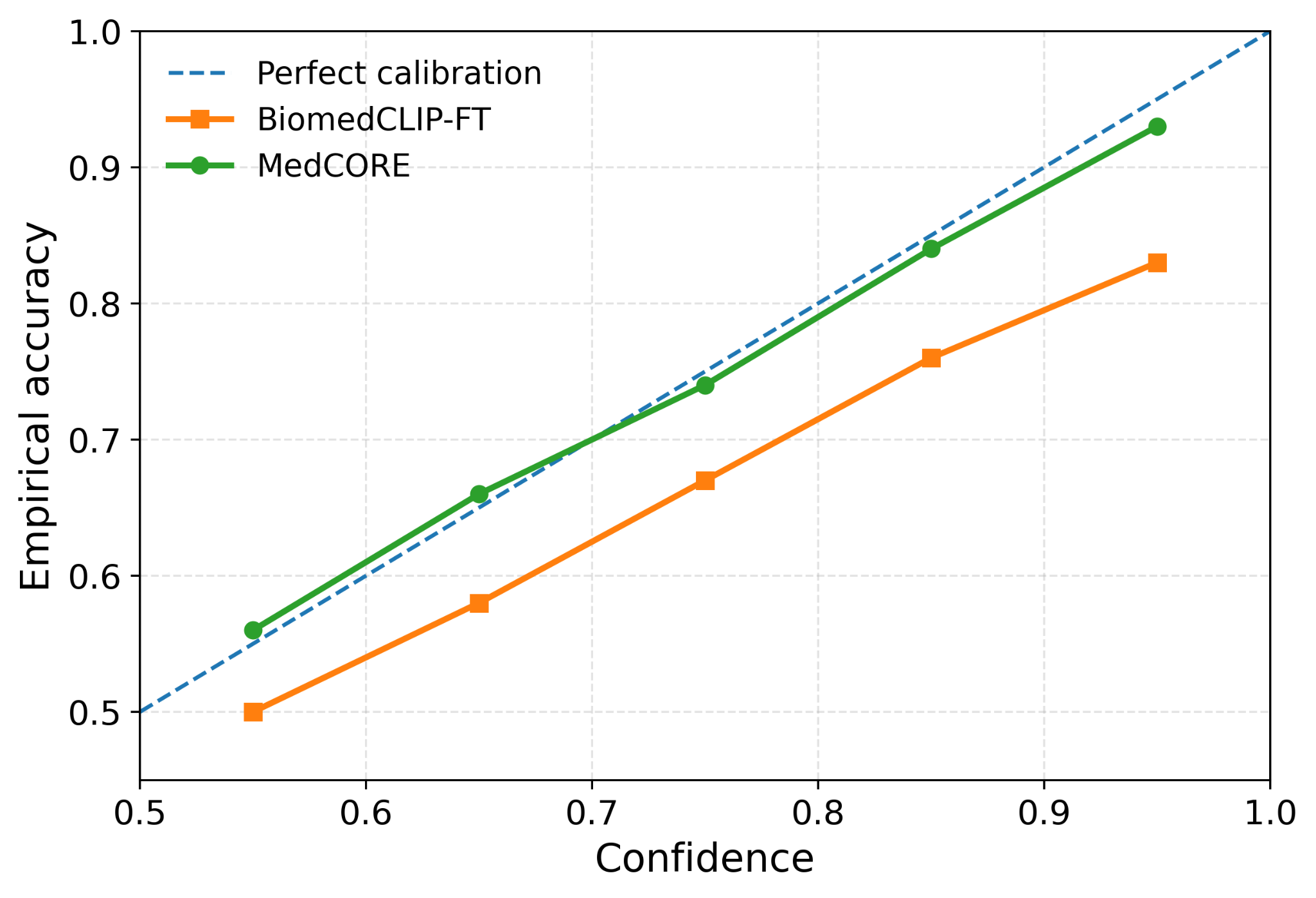}
    \caption{Reliability diagram comparing MedCORE with BiomedCLIP fine-tuning. MedCORE remains closer to the perfect calibration line, indicating better agreement between predicted confidence and empirical accuracy.}
    \label{fig:reliability_given}
\end{figure}

Figure~\ref{fig:reliability_given} compares the reliability diagram of MedCORE with BiomedCLIP fine-tuning. Table~\ref{tab:reliability_values} lists the plotted confidence bins and empirical accuracies. MedCORE follows the diagonal perfect-calibration line more closely than BiomedCLIP-FT across all confidence bins.

\begin{table}[!t]
\centering
\caption{Reliability diagram values from Fig.~\ref{fig:reliability_given}. Values report empirical accuracy at each confidence bin.}
\label{tab:reliability_values}
\vspace{0.3em}
\small
\setlength{\tabcolsep}{5pt}
\begin{tabular*}{\linewidth}{@{\extracolsep{\fill}}cccc@{}}
\toprule
\textbf{Conf.} & \textbf{\shortstack{Perfect\\calibration}} & \textbf{BioMedCLIP-FT} & \textbf{MedCORE} \\
\midrule
0.55 & 0.55 & 0.50 & 0.56 \\
0.65 & 0.65 & 0.58 & 0.66 \\
0.75 & 0.75 & 0.67 & 0.74 \\
0.85 & 0.85 & 0.76 & 0.84 \\
0.95 & 0.95 & 0.83 & 0.93 \\
\bottomrule
\end{tabular*}
\end{table}

BiomedCLIP-FT is consistently below the diagonal, indicating over-confidence: its predicted confidence is higher than its empirical accuracy. This effect is especially visible at high confidence, where the empirical accuracy at 0.95 confidence is only 0.83. In contrast, MedCORE reaches 0.93 empirical accuracy at the same confidence level. Using the plotted bins with equal weighting, the mean absolute calibration gap is approximately 8.2\% for BiomedCLIP-FT and 1.2\% for MedCORE. This improvement is consistent with the uncertainty-calibrated aggregation mechanism, which reduces the influence of unreliable criteria before producing the final prediction.

%----------------------------
\section{Conclusions and Future Directions}

We presented MedCORE, a criteria-grounded framework for interpretable medical image diagnosis. 
Instead of relying on a direct image-to-label prediction pipeline, MedCORE decomposes diagnosis into clinically meaningful criteria, localises criterion-specific evidence, refines inter-criteria relationships with graph attention, and aggregates evidence using uncertainty-calibrated weighting. 
By combining frozen biomedical vision--language representations with lightweight criterion reasoning, visual prototype reinforcement, and text-aligned supervision, MedCORE improves both diagnostic performance and explanatory structure across dermoscopy, breast ultrasound, and retinal fundus imaging benchmarks. 
Experiments on ISIC 2018, BUSI, and IDRiD show that MedCORE achieves stronger accuracy, macro-F1, and AUC than CNN, Transformer, vision--language, concept bottleneck, and prototype-based baselines. 
More importantly, the model provides criterion-level evidence, reliability scores, uncertainty estimates, and spatial heatmaps, making its predictions more transparent and clinically interpretable. 
Future work will extend MedCORE to larger multi-centre datasets, incorporate expert-validated criterion annotations, and study its calibration and usefulness in real clinical decision-support settings.
%----------------------------

\section*{Acknowledgement}
This research was funded by Khalifa University of Science and Technology through the Faculty Startup grant under Project ID: KU-INT-FSU-2026-8471000024.
% \begin{figure}
% \begin{tabular}{ccc}
% \bmvaHangBox{\fbox{\parbox{2.7cm}{~\\[2.8mm]
% \rule{0pt}{1ex}\hspace{2.24mm}\includegraphics[width=2.33cm]{images/eg1_largeprint.png}\\[-0.1pt]}}}&
% \bmvaHangBox{\fbox{\includegraphics[width=2.8cm]{images/eg1_largeprint.png}}}&
% \bmvaHangBox{\fbox{\includegraphics[width=5.6cm]{images/eg1_2up.png}}}\\
% (a)&(b)&(c)
% \end{tabular}
% \caption{Caption Here}
% \label{fig:teaser}
% \end{figure}

%-------------------------------------------------------------------------

%-------------------------------------------------------------------------
\bibliography{egbib}
\end{document}